\pdfoutput=1

\documentclass[11pt]{article}

\usepackage[]{EMNLP2022}

\usepackage{times}
\usepackage{latexsym}
\usepackage{graphicx}
\usepackage{xcolor}
\usepackage{booktabs}

\usepackage[T1]{fontenc}

\usepackage[utf8]{inputenc}

\usepackage{microtype}
\usepackage{acronym} 

\usepackage{inconsolata}
\usepackage{amsfonts}
\usepackage{lipsum}

\title{Bringing the State-of-the-Art to Customers: A Neural Agent Assistant Framework for Customer Service Support }

\author{Stephen Obadinma$^b$ , Faiza Khan Khattak$^a$, Shirley Wang$^{a,c}$, Tania Sidhom$^{a,b}$, \\ \textbf{Elaine Lau$^{d,i}$, Sean Robertson$^{a,c}$, Jingcheng Niu$^{a,c}$, Winnie Au$^a$, Alif Munim$^a$, } \\ \textbf{Karthik Raja K. Bhaskar$^e$, Bencheng Wei$^e$, Iris Ren$^e$, Waqar Muhammad$^e$, Erin Li$^e$,} \\ \textbf{Bukola Ishola$^e$, Michael Wang$^f$, Griffin Tanner$^f$, Yu-Jia Shiah$^g$, Sean X. Zhang$^g$,}  \\ \textbf{Kwesi P. Apponsah$^g$, Kanishk Patel$^h$, Jaswinder Narain$^g$, Deval Pandya$^a$,} \\  \textbf{ Xiaodan Zhu$^{a,b}$, Frank Rudzicz$^{a,c}$, Elham Dolatabadi$^{a,c}$ } \\
  $^a$Vector Institute for Artificial Intelligence, $^b$Queen's University , $^c$University of Toronto, \\ $^d$McGill University, $^e$CIBC, $^f$KPMG, $^g$PwC Canada, $^h$University of Alberta, $^i$Mila\\ \vspace{2cm}   }

\begin{document}
\maketitle


\begin{abstract}
Building Agent Assistants that can help improve customer service support requires inputs from industry users and their customers, as well as knowledge about state-of-the-art Natural Language Processing (NLP) technology. We combine expertise from academia and industry to bridge the gap and build task/domain-specific Neural Agent Assistants (NAA) with three high-level components for: (1) Intent Identification, (2) Context Retrieval, and (3) Response Generation. In this paper, we outline the pipeline of the NAA's core system and also present three case studies in which
three industry partners successfully adapt the framework to find solutions to their unique challenges. Our findings suggest that a collaborative process is instrumental in spurring the development of emerging NLP models for Conversational AI tasks in industry. The full reference implementation code and results are available at \url{https://github.com/VectorInstitute/NAA} 


 




\end{abstract}

\section{Introduction}
\label{intro}
Rising demand for AI-powered conversational agents \cite{fu2022learning, sundar2022multimodal},  especially for customer support service, is estimated to grow at a compound annual growth rate (CAGR) of 23.4\%, earning a predicted revenue of around \$29.9B USD by 2028 \cite{market2018global}. As such, conversational AI research has increased substantially, especially to enhance customer service support \cite{nicolescu2022human}. Despite a proliferation of agent assistants from Microsoft%
, IBM%
, Oracle%
, and Google%
, there remain many unanswered questions and challenges \cite{fu2022learning} that need to be addressed before the widespread proliferation of AI in practice. 
We argue that bridging the gap between natural language processing (NLP) research in academia and industry is an overlooked issue in conversational AI.

Apart from the handful of aforementioned conglomerates, it is difficult for most other companies, which are not in the process of conducting cutting-edge NLP research, to benefit from the recent progress of NLP. Conventional conversational AI architectures are delicate and complex, and require a large degree of specialised knowledge to bring a full system into fruition. 
However, with the introduction of the large-scale pretrained transformer-based language models such as BERT \citep{devlin2018pretraining} and GPT-2 \citep{radford2019language},
it is possible for smaller teams to take advantage of this ``monopoly'' by being able to fine-tune these powerful models on their comparatively small amount of data and achieve high performance, therefore harnessing the architectural achievements of those powerful language models to devise their own conversational AI systems that exploit their more fine-grained expertise and knowledge of their customers' needs.

The interdisciplinary nature of AI-enabled customer service support
makes for an inherently difficulty task \cite{nicolescu2022human}.
To arrive at an answer to a user query, a system must bore through 
several layers of complexity: first, the intent behind the question 
must be categorized and quantized into a form which can be 
manipulated by the system; second, determining from thousands or
millions of stored records the information relevant to that intent;
and finally to manipulate said information into a form which the
user may understand.


To handle such complexity, techniques from across NLP must be employed, raising the barrier-of-entry for
companies with significant customer support needs. Larger,
well-established companies can offer solutions as a service, but
said solutions are often closed-source, trained on generic data
with few in-domain terms, and may not be easily integrated into
a company's workflow.

This paper represents the cumulative efforts of both researchers
and industry practitioners over a yearlong project (whose details may 
be found in Appendix~\ref{sec:appendix}) to develop 
state-of-the-art customer service support systems. In its publication, we 
hope to help lower the barrier-of-entry for future industry 
practitioners by inspiring similar collaborations. Our contributions are twofold and in line with the 
strengths of the contributors:
\begin{enumerate} \itemsep -2mm
    \item We release an open-source Neural Agent Assistant (NAA)
    based on state-of-the-art neural architectures  \cite{fu2022learning}. The
    implementations are well-documented, illustrating how the
    systems can be extended for specific use cases.
    \item To wit, we explore three case studies in which three 
    industry partners successfully adapt the agent to find solutions to their unique challenges.
\end{enumerate}
In their presentation, we stress that a perfect, universal customer 
service system is unattainable as their real-world applications are
not identical. As such, any dialogue about real-world NAA must
feature a discussion of both the technologies shared with, and
the differences between tasks.

\section{Neural Agent Assistant framework for Customer Service Support}
\label{framework}

As the basis of our collaborative project, we established a common Neural Agent Assistant (NAA) framework to serve as a guide for industry partners to adapt to their companies' workflows. Those technologies were made available in a clear, easy-to-use engineering pipeline, alleviating one of the greater challenges for industry in obtaining a foothold in building competent neural agents. 

\begin{figure*}
\centering
\includegraphics[width=0.8\textwidth]{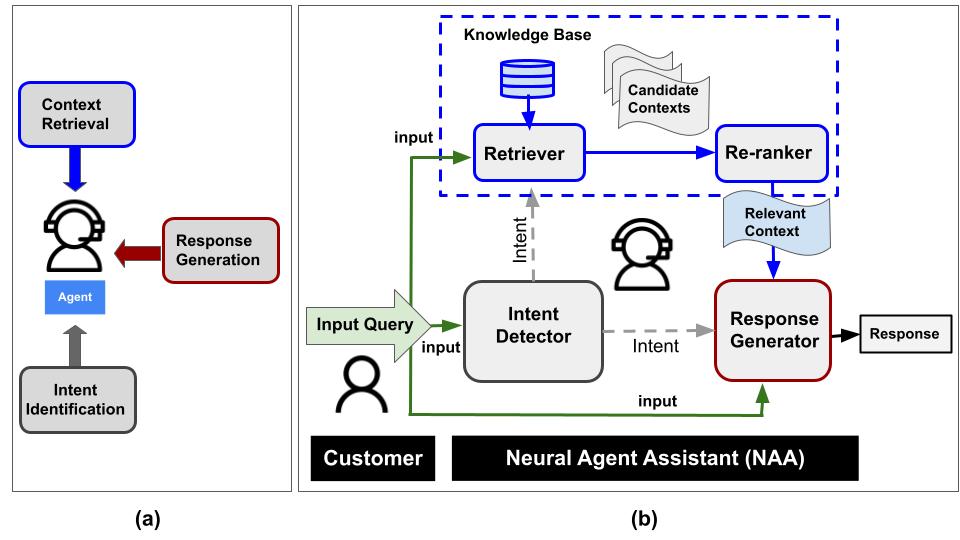}
\caption{Neural Agent Assistant framework. \textbf{(a)} shows how a human agent is able to leverage the three NAA components for support. \textbf{(b)} shows the complete NAA pipeline, starting with an input query from a customer which is first fed to the NAA's intent identification module. The intent label is generated and can be used to guide the context retrieval and response generator. The retriever and re-ranker work together to find the most relevant context candidate passages from the KB, which the response generator employs to craft a response to the user query.}
\label{fig1}
\end{figure*}

The framework features three high-level components shown in Figure~\ref{fig1}: (1) Intent Identification, (2) Context Retrieval, and (3) Response Generation. Though designed as parts in a pipeline in generating a natural language response to customer questions
(Figure~\ref{fig1}~(b)), the output from any given component can nonetheless be transferred back to the human agent when needed (Figure~\ref{fig1}~(a)).

\subsection{Intent Identification}
Intent identification is a critical component of NAA as it is often the first step in a customer service pipeline upon which all subsequent components depend on. Its goal is to classify the intent of a given query. This may be binary, determining whether the query is relevant or irrelevant to the knowledge base (KB) (see Section~\ref{poc}). More often, the task involves classifying the query into a fine-grained, domain-specific intent class. For this purpose, we designed a query encoder to detect and understand customers’ input and classify it into one of $N$ classes. 
The encoder consists of a pre-trained, Transformer-based language model \cite{vaswani2017attention}, followed by a pooling layer, then a final linear layer. A query is fed as an input to the encoder, with the latter outputting a categorical probability distribution over the intents.

We chose BERT \cite{devlin2018pretraining} as the encoder, which we fine-tune and evaluate on two task-specific corpora: Banking77 \cite{Casanueva2020} and CLINC150 \cite{larson-etal-2019-evaluation}. Following the methodology proposed by \cite{zhang2021effectiveness}, we first fine-tune the BERT model for the intent classification task using the Banking77 dataset, which is composed of 13\textit{k} queries labelled into 77 intents. We then evaluate it on the CLINC150 dataset (banking domain) in the few-shot setting using all 15 classes. We achieved 92\% F1-score for fine-tuning and 85\% and 90\% F1-score for one-shot and five-shot learning, respectively.

For NAA's core system demonstration, the intent label is not used to guide the context retrieval and response generation, but it can be easily leveraged for knowledge extraction and abstractive summarization as shown in Section 3.

It is important to note that intent identification is only one of the main natural language understanding (NLU) components that a complete customer support agent would possess. Named-entity recognition (NER) is often an equally important task, where the goal is to identify any named entities, including relevant people, organizations, or products the user is inquiring about, as well as details about time and location, like for scheduling a restaurant reservation. There can be a great need for an agent to identify domain-specific entities in user dialog for the purposes of slot filling, hence there is a similar need for NER frameworks for easily adapting to domain-specific industry needs. For the sake of simplicity, and because an NER component was not requested from our industry partners, it is not included within the scope of the project. We leave developing this NER framework to future work.

\begin{table}
\centering
\caption{Primary results for Passage Retriever (PR) and the Re-Ranker (RR), comparing the performance of pre-trained and fine-tuned models with a variety of standard passage retrieval evaluation metrics (note higher results are better).}
\label{contextretrievalresults}
\resizebox{0.9\linewidth}{!}{%
\begin{tabular}{llll} 
\hline
 &  & MRR & MAP \\ 
\hline
\textbf{PR} & \textbf{pretrained~} & \multicolumn{1}{r}{} & \multicolumn{1}{r}{} \\
 & msmarco-distilbert-base-tas-b & 0.835 & 0.801 \\
 & \textbf{fine-tuned~} &  &  \\
 & multi-qa-mpnet-base-dot-v1 & \multicolumn{1}{r}{\textbf{0.886}} & \multicolumn{1}{r}{\textbf{0.860}} \\
 & msmarco-distilbert-base-tas-b & \multicolumn{1}{r}{0.864} & \multicolumn{1}{r}{0.834} \\ 
\hline
\textbf{RR} & \textbf{pretrained~} &  &  \\
 & msmarco-MiniLM-L-6-v2 & \multicolumn{1}{r}{0.998} & \multicolumn{1}{r}{0.997} \\
 & \textbf{fine-tuned} &  &  \\
 & msmarco-MiniLM-L-6-v2 & \multicolumn{1}{r}{\textbf{1.000}} & \multicolumn{1}{r}{\textbf{1.000}} \\
\hline
\end{tabular}
}
\end{table}

\subsection{Context Retrieval}
This component is designed to retrieve the most semantically similar documents from a predefined knowledge base (corpus of documents) given a user query. Documents or
passages extracted from the KB are presumed to contain an appropriate answer to the query, though not in a concise form. Context retrieval is a two-step process starting with passage retrieval and followed by re-ranking. The retriever uses a bi-encoder design \cite{karpukhin-etal-2020-dense}, extracting supporting context from a knowledge-base at the level of paragraphs (hence asymmetric semantic search) based on SentenceBERT \cite{reimers-2019-sentence-bert}. The re-ranker is a BERT-based encoder designed for sequence-pair classification that scores the relevancy of all top-k retrieved candidate contexts for a given input query \cite{2019arXiv190104085N}.\\
\indent Our implementation includes pipelines to run the retriever and re-ranker pre-trained on MSMARCO \cite{bajaj2016ms} as-is, or to fine-tune both models using Multiple Negatives Ranking Loss on a knowledge-grounded, question-answering corpus such as ELI5 \cite{fan-etal-2019-eli5}. The MSMARCO dataset was generated by sampling and anonymizing Bing usage logs. The dataset includes over 1 million queries with at least one human-generated answer per query, as well as relevant Wikipedia passages retrieved by Bing for each query. The ELI5 dataset is made up of complex questions and long and explanatory answers from Reddit users about random topics. Each question and answer pair is grounded in relevant Wikipedia passages for supporting information. 
Table~\ref{contextretrievalresults} compares and reports the performance of the pretrained and fine-tuned models on the ELI5 dataset, illustrating the benefits of fine-tuning across systems
and conditions.

Different alternative approaches have been taken in previous works for context retrieval, including combining dense passage retrieval with lexical rule-based retrieval such as BM25 to improve results and computational efficiency \cite{gao-etal-2021-coil, khazaeli-etal-2021-free}, and using extensive pretraining on the encoders \cite{Khattab2020ColBERTEA}. Although these methods have been shown to be effective in some cases, it adds further complexity and makes for a less generalizable approach compared to the BERT-based approach we take. 

\subsection{Response Generation}
The ultimate goal of our NAA is to be able to automatically generate a human-like answer to a customer's query. Therefore, the final component is dedicated to producing in-context natural-language responses to customers' queries via a generative transformer model. The model is fed the ranked passages from the retrieval component and outputs the natural-language response. Though the input is extractive, the output is abstractive.

Our reference implementation provides pipelines for training GPT-2 models of any size \cite{radford2019language} for the question answering (QA) task on both of the above-mentioned knowledge-grounded datasets: MSMARCO and ELI5. 
We used a multi-task loss combining language modeling with a next-sentence prediction objective. Following training, nucleus top-p sampling was used for decoding and text generation. The results for GPT-2 medium are presented in Table~\ref{tblgpt}.

\begin{table}[]
\caption{Primary results for the Response Generation component of NAA. Models were fine-tuned on MSMARCO.}
\resizebox{1\linewidth}{!}{%
\begin{tabular}{@{}lllll@{}}
\toprule
\hline
\multicolumn{1}{c}{\textbf{Model}} & \multicolumn{1}{c}{\textbf{Test-set}} & \multicolumn{1}{c}{\textbf{F1-score}} & \multicolumn{1}{c}{\textbf{BLEU-1}} & \multicolumn{1}{c}{\textbf{Rouge-L}} \\ 
\hline
\midrule
GPT2-meduim                        & MSMARCO                               & 40.2\%                          & 32.0\%                            & 36.0\%                               \\
GPT2-large                         & MSMARCO                               & 32.0\%                          & 20.4\%                            & 28.0\%                               \\
DialoGPT-medium                    & MSMARCO                               & 28.2\%                          & 15.3\%                            & 24.6\%                               \\
GPT2-meduim                        & ELI5                                  & 6.98\%                          & 0.1\%                             & 5.22\%                               \\ \bottomrule
\hline
\end{tabular}
}
\label{tblgpt}
\end{table}





The three components of our pipeline - intent classification,
context retrieval, and response generation - perform well
on the domains and tasks they were trained on. However, as discussed
in Section~\ref{intro}, a general system will fail to account for
the specific challenges companies face related to the
content and quality of NAA. As was illustrated, the pipeline
above provides numerous opportunities for tailoring each function,
opportunities leveraged by our industry partners in the next 
section.

\section{Industry-specific Proof Of Concept Implementations}
\label{poc}

In this section, we demonstrate how the NAA framework outlined
in Section~\ref{framework} was adapted to our industry partners' use-cases and settings, which is continuing to be refined for deployment in real-world scenarios.


Canadian Imperial Bank of Commerce (CIBC) is a financial institution with an Advanced Analytics team focusing on finding the correct curated response to banking queries. KPMG is a professional services firm which is looking to NAA to service internal queries over large bodies of legal and regulatory documents. PricewaterhouseCoopers (PwC) is also a professional services firm looking to better advise its banking clients on building NAA technologies.
In the following subsections we clarify the companies' motivations
in participating in the project, how they adapted our NAA framework,
and their plans for deployment.




\subsection{Company I: CIBC - NAA Tools for Banking Customer Service Support}
The primary motivation for implementing and deploying NAA by CIBC is to support the bank’s continued focus on leveraging digital technologies to make clients’ banking experiences even better.

The pipeline modified by CIBC (see Figure \ref{cibc_fig_1} in Appendix \ref{sec:appendix}) consisted of four components. The first component was a binary classifier (i.e., a BERT encoder) which was trained to classify whether an input query has a banking intent. The inputs identified as being banking related were further processed by the assistant. Module 2 was the same as NAA's intent identification component, though it was improved through data augmentation: the component was fine-tuned on back translation \cite{sennrich-etal-2016-improving_bt} (English $\rightarrow$ German $\rightarrow$ English) and insertion data. With data augmentation, F1-scores improved from 91.8\% to 92.7\%. Module 3 was a knowledge-driven QA system including both the Context Retrieval and Response Generation components of the NAA. If the banking intent from the previous layer had a confidence score greater than parity, then the top 5 most relevant financial contexts from a curated financial KB were retrieved and ranked. The KB included long-format questions and answers, as well as FAQs downloaded from the company's publicly available website. The context, therefore, was question/answer pairs that, after being retrieved, provided a similar question along with an answer to the customer query. In addition, the GPT-2 Medium model pretrained on MSMARCO was also used to generate human-like responses for the questions based on the retrieved long answers. Low-confidence queries were passed into the final component of the pipeline, which is based on KeyBERT, an architecture utilizing BERT, and deals with out-of-domain intent identification. Using the KeyBERT, keywords from the query were extracted and kept as new out-of-domain intents. Moreover, these new out-of-domain intents were recorded and tracked to enhance banking intent classification in the future by creating more relevant intent categories.\\
\indent Following a successful implementation of the pipeline, a web application was developed to be used directly by clients. React and REST API service were used as the basis for the web application, which were integrated with the Amazon Web Services (AWS) cloud platform. Extensive developer testing has been conducted, and additional deployment include carrying out agent volunteer testing, which involves using several agents at the bank testing the application for optimal functionality and usability. Lastly, in order to make the system able to dynamically improve in response to customer feedback, the developers plan to implement a feedback loop using customer data. The feedback loop is a mechanism that helps determine how well the model works in production, and provides the necessary feedback to determine whether any changes are needed as a response to customers' user experience with the application and any desired improvements (see Figure \ref{cibc_fig_2} in the Appendix).

\subsection{Company II: KPMG - A Q\&A Tool for Legal Documents Analysis}

\begin{table}
\centering
\caption{KPMG: Legal data NAA pipeline results. Passage size (PS) refers to number of sentences used for the passages (2, 3, and 4 for \textit{short}, \textit{medium}, and \textit{long} respectively). Corpus refers to whether the dataset was preprocessed. Retriever refers to corpus embedding bi-encoder (all retrieval components are pretrained on MSMARCO). Response generator (RG) shows the pretraining dataset used for GPT-2-medium.}
\label{kpmg_results}

\resizebox{\linewidth}{!}{%
\begin{tabular}{llllrrr} 
\hline
\textbf{PS} & \textbf{Corpus} & \textbf{Retreiver} & \textbf{RG} & \multicolumn{1}{l}{\textbf{Manual Score}} & \multicolumn{1}{l}{\textbf{BLEU-1}} & \multicolumn{1}{l}{\textbf{F1-score}} \\ 
\hline
Med. & Raw & DISTILBERT & ELI5 & 44 & 0.193 & 0.218 \\
Med. & Clean & DISTILBERT & ELI5 & 40 & 0.176 & 0.198 \\
Short & Clean & DISTILBERT & ELI5 & 30 & 0.181 & 0.206 \\
Long & Clean & DISTILBERT & ELI5 & 43 & 0.212 & 0.232 \\
Med. & Clean & BERT & ELI5 & 49 & 0.303 & 0.336 \\
Med. & Raw & BERT & MSMARCO & \textbf{74} & 0.450 & \textbf{0.506} \\
Med. & Clean & BERT & MSMARCO & \multicolumn{1}{r}{N/A} & \textbf{0.459} & 0.443 \\
Short & Clean & BERT & MSMARCO & \multicolumn{1}{r}{N/A} & 0.422 & 0.426 \\
\hline
\end{tabular}
}
\end{table}

The motivation for adapting NAA and its implementation by KPMG was reduce the inefficiencies in providing accurate information to user queries in the legal domain drawn from agreements and reports. The company aims to improve their existing text search parsing relying on manual/keyword search using a more robust and easy-to-use novel knowledge-based document query tool based on a NAA.  

The pipeline for this implementation included using and externally evaluating the context retrieval and response generation components of NAA framework on a curated legal dataset, CUAD v1 \cite{hendrycks2021cuad}, which contains over 13\textit{k} sentences based on commercial legal contracts labelled into 41 types of legal clauses. A group of 5 experts including 3 data scientists and 2 legal domain data engineers helped create the legal domain knowledge grounded QA dataset and provided manual evaluation of the performance of the model on the response generation task. The legal QA dataset
includes in-house human generated question and answer pairs (n=47) drawn from CUAD v1. In particular, the dataset is a collection of question-answer-evidence triplets,
and a KB where the questions and answers were generated by the experts and were accompanied by a supporting context from the KB. The quality and accuracy of the generated responses by the pipeline was evaluated manually by one of the data engineers with legal domain expertise, as well as automatically by using BLEU and F1-scores as can be seen in Table \ref{kpmg_results}. Through manual evaluation, each generated response was scored from 0 to 100 based on their quality and relevance, and the average score of all of them was calculated.  The retriever and GPT-2-medium fine-tuned on MSMARCO achieved the highest score of 74 through manual evaluation, as well as a BLEU-1 and F1-score of 0.45 and 0.5, respectively.


A two-phase production deployment plan is implemented for the developed prototype. In phase I, the prototype is planned to be made available to internal employees only as a document processing tool. The service will be based on a cloud-hosted managed service environment, and would allow users to upload a set of documents and enter queries. Based on the queries, related context will be extracted from the documents and abstractive responses, which can be used in the formation of summary reports. Built on phase I, in phase II the tool will be made available to external clients and integrated with their systems through an API for the their managed services platform.


\subsection{Company III: PwC Canada - Transferring Emerging Technologies to Financial Customer Queries}


The main motivation for adapting and applying NAA by PwC was to gain experience in creating an end-to-end NAA pipeline to process customer queries regarding bank account opening and mortgage applications. The aim was to later use the acquired knowledge to support small to mid-sized financial clients in building NAA chatbots that can reduce inefficiencies when dealing with large volumes of requests (particularly during high-volume seasons like home-buying seasons).

The first component of the prototype included a binary classifier to classify the text into either a “general” or “agent-related” topic to mitigate the answer generation model from confusing the user by responding to irrelevant queries. The binary classifier was trained on a labeled dataset including a total of 15,732 queries (10,331 ``agent-related'' and 5,401 ``general''), and achieves an F1-score of greater than 0.95. The ``general'' queries were extracted from English-Second-Language practice conversations and common greetings, while ``agent-related'' queries were pulled from the Banking77 dataset \cite{Casanueva2020}. In case of a general query, a pretrained dialogue generation model, DialoGPT \cite{zhang2019dialogpt}, was triggered to make a basic response to the user. 

The answer generation component included both the retrieval
and response generation tasks. For the retrieval task, both the retriever and re-ranker provided by NAA (pretrained on  MSMARCO) were used to extract relevant question-answer pairs from the KB. The KB included 250 long format questions and their respective answers downloaded from FAQs owned by top banks in Canada. Following the retrieval task, the top 3 contexts (including question-answer pairs similar to section 3.1) along with the user's query were provided as inputs to the answer generation model which uses a GPT-2 model further fine-tuned on the 250 question answer sets. The generated answers were compared and evaluated on a set of new human-generated question answer pairs (n=30), achieving a Rouge1 F1-score of 0.088 and a RougeL of 0.084.

The deployment of this solution at the production level is to combine NAA components with Automated Speech Recognition (ASR) tools using a backend service that determines whether a client request is audio or text, after which the request would be processed by the appropriate pipeline. An asynchronous web framework such as NodeJS’ Express or Python’s FastAPI is planned to be used to create a system scalable to multiple concurrent users, and using an Apache MXNet framework in combination with distributing the workload across multiple GPUs will support inference on the ASR and the response generation models.

\section{Summary and Discussion}
\label{lessons}

We summarize our experiences in this section.

\vspace{-2mm}
\paragraph{Domain/task-specific system works better:} As stated earlier, the task/domain agnostic system is not well-suited for different use-cases as our industry partners found, and as confirmed
 by our experiences. 
     Real-world conversational datasets are noisy and contain domain-specific concepts, and as a result, domain agnostic systems that are trained on clean open-source datasets are likely to have poor performance when evaluated on realistic inputs. The best practice would be to have a system designed/adapted based on domain and task. 
\paragraph{Finding a good performance evaluation method:} Due to the inherent difficulty in evaluating the quality of retrieved contexts and generated answers, finding a suitable model assessment strategy is imperative \cite{khapra2021tutorial}, which includes the processes of model selection, bench-marking, and deciding between manual and automated assessment. Automatic evaluation metrics in particular, although flawed, allow for the ability to benchmark many different model variations, leading to a strong model being picked if there is a general agreement between different metrics. The companies found that using a combination of automatic evaluation metrics like BLEU score, F1-score, Rouge-1, along with expert manual evaluation, helped better inform model selection.
\vspace{-2mm}
\paragraph{Efficiency and optimization:} The efficiency of the algorithm/model being used is important, as is weighing trade-offs and discovering optimizations. For example it was noted that BERT-base models tend to perform better than DistilBERT-base for the bi-encoder, but at the cost of longer inference time. Hence, it is important to consider multiple different approaches when implementing a component and deciding what is the most important aspect of performance. Another challenge that was identified by one company was designing an efficient algorithm for parsing customer inputs into meaningful entities, which was solved by using a custom built parser tailored specifically for in-domain data.
\vspace{-2mm}
\paragraph{Effectively Combining Academia and Industry Expertise:}
This collaborative project combined the expertise from academia and industry, which resulted in successful implementation of systems to be deployed at the industry participants' companies. We tried to keep the whole process smooth by having regular meetings and adding documentation. This helped us to collaborate in an efficient and effective manner while keeping engagement throughout the whole process.

\section{Conclusion}
\label{conclusion}
In this paper, we present a collaborative project for building three customized NAA systems for different tasks and domains in industry. Through our framework and collaborative process, we have tried to bridge the gap between industry and cutting edge NAA technologies by providing much needed open-source tools and expertise in foundational language models and conversational AI. We have observed this framework being successfully adapted into real-world use-cases to enhance customer service support, and it has helped spur NAA development in industry. We hope that our current effort will serve as a valuable use-case to the community, and it is our plan to continue similar collaborative efforts in future in other areas of AI.

\section{Ethical Considerations}
\label{considerations}
A limitation of the natural language response generation module of our framework is that there are few guarantees that the model does not produce erroneous or harmful responses in response to user queries. As the base of this module is a language model such as GPT-2 that is pretrained on a massive corpus of data that has not been carefully ensured to be debiased and free from harmful or prejudiced content, it can be susceptible to adversarial inputs and thus has the potential to produce bigoted responses even to benign user queries \cite{kirk2021bias}. This presents ethical concerns when having the NAA deal directly with customers, and therefore, to avoid such concerns, we recommend that the NAA in its current form only be used in a human-in-the-loop process, where a human agent works closely with the NAA for assistance. The human agent can ensure that any potential customer or client does not receive any unwanted responses from the NAA.

\section*{Acknowledgements}
We would like to acknowledge and give thanks to Sedef Akinli Kocak and Gerald Penn from the Vector Institute for Artificial Intelligence for their help in facilitating this project. We also acknowledge the contributions of Andrew Brown from CIBC on CIBC's NAA implementation. 

\bibliography{main}
\bibliographystyle{acl_natbib}

\appendix

\newpage
\section{Appendix}
\label{sec:appendix}

\textbf{Project participants:} The participants of this project consist of:
\begin{itemize}
    \item  \textit{Technical and project management teams from Vector Institute for AI} initiating this project with the mission to bring industry and academia closer so that the two groups could learn and benefit from each other.
    \item \textit{Data scientists from three industry partners} participated to enhance their knowledge and apply new methods in their companies.
    \item \textit{University faculty from  partnering universities} providing additional advising and exposure to the state-of-the-art methods. 
\end{itemize}

\textbf{Project steps:} The following are the points  describing the project steps and participation.
\vspace{-2.5mm}
\begin{itemize} \itemsep -0.9mm    
    \item Vector Institute for AI: 
    \begin{itemize}
        \item  defined the problem statement and project scope with the help of some input from the industry partners,
        \item implemented general domain-agnostic NAA reference implementations consisting of three high-level components: (1) Intent Identification, (2) Context Retrieval, and (3) Response Generation,
        \item  provided tutorial, open-source datasets, computing services, training, and constant feedback \& support.
    \end{itemize}
   
    \item Faculty provided advising on cutting edge technologies.
    \item Industry partners (CIBC, KPMG, PwC):
\begin{itemize}
    \item  identified the use-cases,
    \item  modified, trained/re-trained and improvised the framework on the datasets of their choice according to their use-cases and industry settings. 
    \item  developed their own domain/industry specific end-to-end pipeline to deploy in their respective companies.
\end{itemize}
    
\end{itemize}


\section{Environment Setup}
The following resources were provided by the \textit{Vector Institute for AI} to the participants.

\begin{itemize}
    \item \textbf{Git repos:} Git repositories containing full reference implementation code and training details were provided by the \textit{Vector Institute for AI}\footnote{\url{https://github.com/VectorInstitute/NAA}}.
    \item \textbf{Cluster and GPU access:} Access to cluster and GPUs was provided by the \textit{Vector Institute for AI}. Model training was conducted on a remote cluster on one of 3 GPUs: NVIDIA T4 (16GB), NVIDIA® Tesla P100 (16GB), and NVIDIA RTX 6000 GPU (24GB)
    \item \textbf{Dataset storage:} All datasets used were made available on cluster. Moreover, each participant had  50GB storage space (which could be increased upon request) to store datasets, model checkpoints, and other files necessary to formulate their solutions.
    \item \textbf{Google Cloud Platform (GCP) access:} The participants were provided access to GCP and TPUs. Also a GCP training session from Google was hosted by the \textit{Vector Institute for AI}. The purpose of this platform was to provide the participants with experience using cloud services to train and deploy their system.
    
\end{itemize}

\section{Reference Implementation Training Details}
In this section, we provide some of the pre-processing decisions, training details, and hyperparameters for our reference implementations for replication purposes. For our implementation of the transformer-based language models we use the python libraries, Huggingface Transformers \cite{huggingface_transformers} for BERT and GPT-2, and SentenceTransformers (SBERT) \cite{reimers-2019-sentence-bert} for the bi-encoder and cross-encoder. We use Pytorch as the backbone of our implementations. An AdamW optimizer is used for the training of all of the models.

\begin{itemize}
    \item \textbf{Intent Identification}: The max sequence length for BERT is set to be 300 tokens. The model is fine-tuned for 40 epochs using a batch size of 16 and a learning rate of 2e-5. We utilize a linear scheduler with a warmup ratio of 20\%. We used 5\% of the training set for model validation purposes, and the provided Banking77 test set for evaluation. Early stopping based on validation loss with a patience of 2 was used to select the best performing model. The training settings for the few shot model include using a frozen BERT encoder previously fine-tuned on Banking77 as a feature extractor for few shot classification. A new linear layer on top of the BERT encoder is trained to classify the new 15 classes from CLINC150. The linear layer is trained for 5 epochs with a batch size of 6 and a learning rate of 0.001. The rest of the training settings are similar to before.
    \item \textbf{Context Retrieval:} For the version of the bi-encoder fine-tuned on the ELI5 dataset, the dataset is preprocessed such that a re-ranker pretrained on MSMARCO re-ranks the original 7 wikipedia passages per query, and picks the most relevant passage to be part of the question-answer training pair. The base \textit{msmarco-distilbert-base-tas-b} bi-encoder is fine-tuned on this dataset for 3 epochs with a batch size of 16. A scheduler with a warmup ratio of 10\% is also used.  15\% of the training data is set aside for testing. The cross encoder has the same hyperparameters, with the addition of the max sequence length being set to 512 tokens. A learning rate of 2e-5 is used for fine-tuning both modules.  
    \item \textbf{Answer Generation:} Lastly, the GPT-2 answer generation model generates an output sequence that is a maximum of 200 tokens, at a temperature of 0.7, and  with top\_k and top\_p values of 100 and 0 respectively. The model is trained for 5 epochs on MSMARCO using a batch size of 1 (with 8 gradient accumulation steps simulating a batch size of 8), and a learning rate of 5e-5. Furthermore, the maximum number of input tokens is set to 330. Additional settings include setting the language modelling loss coefficient to 10.0, applying gradient clipping with a magnitude of 10.0, and using a multiple choice classification head for the second head of GPT-2 with a loss coefficient of 1.0. The MSMARCO dataset is preprocessed such that we use the well-formed answer if possible, and avoiding using questions for which there are multiple answers without a well formed answer. We use the existing train and validation sets for fine-tuning and evaluating the model. 
    
\end{itemize}

\clearpage

\onecolumn

\section{Supplementary Figures}

\begin{figure}[!htb]
\centering
\includegraphics[width=0.8\textwidth]{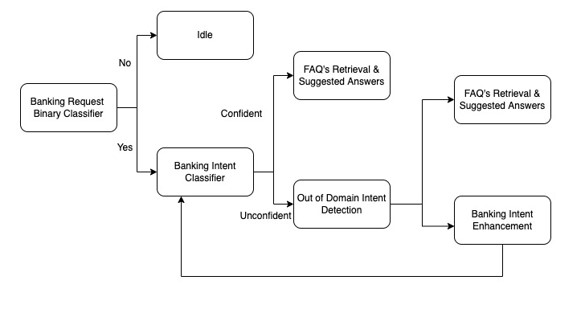}
\caption{CIBC Banking Agent Assistance Implementation Pipeline.}
\label{cibc_fig_1}
\end{figure}

\begin{figure}[!htb]
\centering
\includegraphics[width=0.7\textwidth]{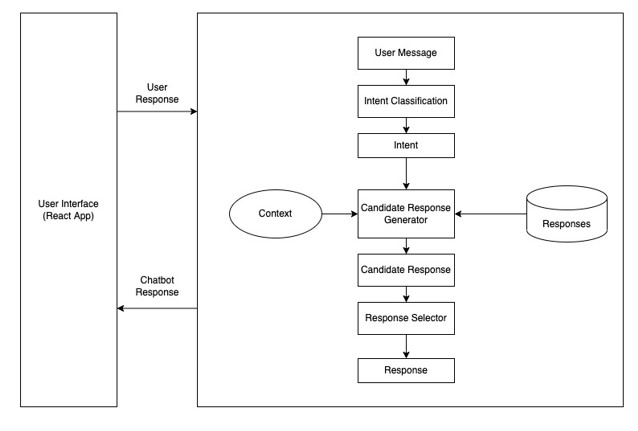}
\caption{CIBC - Diagram showing the Banking Chatbot Service API. The user (human agent) interacts with the NAA pipeline through the React App, and sends a response. Then the NAA pipeline processes the request and a chatbot sends a response.}
\label{cibc_fig_2}
\end{figure}

\end{document}